\documentclass[a4paper,11pt,twocolumn,twoside]{article}
\usepackage{graphicx}
\usepackage{tipa}
\usepackage{sepln}
\usepackage{fullname}
\usepackage[spanish,es-nosectiondot, es-tabla, es-noindentfirst, es-nolists]{babel}
\usepackage[anythingbreaks]{breakurl}
\usepackage[ruled,vlined,linesnumbered]{algorithm2e}
\usepackage[noend]{algpseudocode}
\usepackage{covington}
\usepackage{booktabs}

\usepackage[]{graphics}

\usepackage{epsfig}
\usepackage{epstopdf}

\usepackage[utf8]{inputenc}
\input epsf

\title{Escansión automática de poesía española sin silabación}

\author {\textbf{Guillermo Marco Remón$^1$}, \textbf{Julio Gonzalo}$^1$ \\
	$^1$nlp.uned.es, Research Group in NLP \& IR, UNED, Madrid, Spain,\\
gmarco, julio\{@lsi.uned.es\}}

\seplntranstitle{Automatic Scansion of Spanish Poetry without Syllabification}

\seplnclave{patrones métricos, escansión automática.}

\seplnresumen{En los últimos años, han surgido diversas herramientas de análisis métrico automático de poesía española. Estos sistemas se basan en complejos métodos de silabación y asignación de acentos, los cuales se apoyan en librerías de etiquetado gramatical, cuyo coste computacional es elevado. Este coste incrementa con el cálculo de ambigüedades métricas. Además, no consideran asuntos determinantes en el cómputo silábico como los fenómenos de compensación entre hemistiquios de versos de más de once sílabas. No obstante, es posible llevar a cabo un análisis métrico informativo y preciso sin utilizar estos métodos costosos. En este trabajo se propone un algoritmo que realiza escansiones (número de sílabas, patrón métrico y tipo de verso) sin silabación. El algoritmo resuelve ambigüedades métricas y tiene en cuenta la compensación hemistiquial. Nuestros resultados indican que obtiene una mejora relativa sobre el actual estado del arte de un 2\% en la clasificación de patrones métricos en poesía de medida fija y un 25\% en poesía polimétrica. También se ejecuta 21 y 25 veces más rápido, respectivamente. Por último, se ofrece una aplicación de escritorio como herramienta para los investigadores de la poesía española.}

\seplnkey{metrical patterns, automated scansion.}

\seplnabstract{In recent years, several systems of automated metric analysis of Spanish poetry have emerged. These systems rely on complex methods of syllabification and stress assignment, which use PoS-tagging libraries, whose computational cost is high. This cost increases with the calculation of metric ambiguities. Furthermore, they do not consider determining issues in syllabic count such as the phenomena of compensation between hemistichs of verses of more than eleven syllables. However, it is possible to carry out an informative and accurate metric analysis without using these costly methods. We propose an algorithm that performs accurate scansion (number of syllables, stress pattern and type of verse) without syllabification. It addresses metric ambiguities and takes into account the hemistichs compensation. Our algorithm outperforms the current state of the art by 2\% in fixed-metre poetry, and 25\% in mixed-metre poetry. It also runs 21 and 25 times faster, respectively. Finally, a desktop application is offered as a tool for researchers of Spanish poetry.}

\firstpageno{1}

\begin{document}
	
	
	\setlength\titlebox{19cm} 

	\label{firstpage} \maketitle
	
	%
	\section{Introducción}
	En los últimos años, han surgido diversas herramientas de análisis métrico de poesía española. Dado un verso, extraen automáticamente el número de sílabas y su ritmo. Por ejemplo, a partir del verso:
	\begin{subexamples}\label{example.3}
		\textit{Amigos, el amor me perjudica} \\
		\textit{A-\textbf{mi}-gos-el-a-\textbf{mor}-me-per-ju-\textbf{di}-ca} \\
		11 | 2.6.10 \\
		(Julio Martínez Mesanza) \\
	\end{subexamples}
	se determina que es un verso de once sílabas y su ritmo (también llamado esquema acentual o patrón métrico) es 2.6.10
	
	Para resolver este problema, los sistemas publicados hasta el momento se basan en complejos métodos de silabación y asignación de acentos, los cuales se apoyan en librerías de etiquetado gramatical, cuyo coste computacional es elevado. Como la escasión de un verso puede ser ambigua --los versos del Ejemplo \ref{example:claudio} se pueden escandir a la manera (a) o a la manera (b)--, este coste incrementa con el cálculo de las ambigüedades métricas.
	
	\begin{subexamples}[preamble={\textit{dentro de su fluir los manantiales}}]\label{example:claudio}
		\item\label{example:claudioa}
		\textit{\textbf{den}-tro-de-su-\textbf{fluir}-los-ma-nan-\textbf{tia}-les
		}\\
		10 | 1.5.9\\
		\item\label{example:claudiob}
		\textit{\textbf{den}-tro-de-su-flu-\textbf{ir}-los-ma-nan-\textbf{tia}-les} \\
		11 | 1,6,10 \\
		(Claudio Rodriguez)
	\end{subexamples}
	
	Los versos por encima de las once sílabas se dividen en semiversos o hemistiquios para su escansión. Esta separación introduce fenómenos métricos que afectan al cómputo silábico, y por ello su consideración es fundamental para la correcta determinación del ritmo.  Los métodos actuales no consideran estos fenómenos.
	
	Este trabajo parte de la hipótesis de que es posible llevar a cabo un análisis métrico informativo y preciso sin utilizar métodos de silabación ni librerías de etiquetado gramatical. 
	En la Sección \ref{sec:definicion} comenzaremos definiendo el problema de la escansión e introduciendo el marco teórico necesario para abordar su solución, y en la Sección siguiente resumiremos el estado del arte. 
	
	A partir del conocimiento métrico expuesto, en la Sección \ref{sec:propuesta} se desarrolla el sistema análisis métrico automático que simplifica el problema de la medida del verso gracias a que prescinde de realizar  silabación. El algoritmo tiene en cuenta la compensación hemistiquial y resuelve las ambigüedades métricas derivadas de sinalefas, dialefas, sinéresis y diéresis, sin perder precisión ni información sobre la decisión del analizador. La salida de este método ofrece el número de sílabas, el esquema métrico, el esquema métrico sin acentos extrarrítmicos, su nombre, su grado de coincidencia con el esquema métrico sin acentos extrarrítmicos, así como la forma del verso que ha llevado al sistema a clasificarlo en ese tipo: marca sinalefas, dialefas, sinéresis y diéresis.
	
	En la Sección \ref{sec:evaluación} se evalúa el sistema sobre conjuntos de datos de poesía de metro fijo y mixto. En nuestra experimentación, el algoritmo obtiene una mejora sobre el actual estado del arte de un 2\% en la clasificación de patrones métricos sobre poesía de medida fija y de un 25\% sobre poesía polimétrica. También se ejecuta 21 y 25 veces más rápido, respectivamente. Al final de la sección se analizan los errores cometidos por el algoritmo.
	
	Por último, en las Secciones \ref{sec:app} y \ref{sec:conclusiones} se describe {\em Jumper} una aplicación de escritorio que puede servir como herramienta para los investigadores de la poesía española, y se extraen conclusiones.

	\section{Definición del problema}
	\label{sec:definicion}
	Un verso es una serie de sílabas acentuadas y no acentuadas delimitada por pausas métricas \cite{caparros2001diccionario}. La distribución de los acentos sobre esta serie determina el patrón métrico. Así, la escansión consiste en clasificar los versos por su medida (número de sílabas) y su patrón métrico. 
	
	La sílaba es la unidad estructural de la palabra. En español, el núcleo de una sílaba es siempre vocálico \cite[8.1a]{espanola2009nuevafonetica}. No obstante, la vocal también desempeña funciones de margen silábico en los diptongos; se forman combinando las vocales abiertas con cerradas, cerradas con abiertas, y las cerradas entre sí. Son un conjunto de 14 realizaciones: \textit{ai, au, ei, eu, oi, ou, ui, iu, ia, ua, ie, ue, io, uo}. 
	
	En general, las palabras en español tienen sólo un acento, con algunas excepciones como los adverbios acabados en \textit{-mente}. En función de dónde recae su acento, se clasifican y acentúan gráficamente de la siguiente manera:
	
	\begin{itemize}
		\item Palabras oxítonas o agudas: cuando la sílaba acentuada es la última de la palabra. Aunque existen algunas normas adicionales \cite[3.4.1.2.1]{espanola2010ortografia},  principalmente se acentúan con un signo gráfico cuando terminan en los grafemas de \textit{n}, \textit{s} o \textit{vocal}.
		\item Paroxítonas o llanas: cuando la sílaba acentuada es la penúltima de la palabra. Se acentúan con un signo gráfico cuando terminan en un grafema distinto de \textit{n}, \textit{s} o \textit{vocal}.
		\item Proparoxítonas o esdrújulas: cuando la sílaba acentuada es la antepenúltima de la palabra. Se acentúan con un signo gráfico siempre.
	\end{itemize}
	
	De manera homónima, un verso, por la posición de su última sílaba acentuada, recibe el nombre de oxítono, paroxítono o proparoxítono. De cada uno de ellos, resultan unos fenómenos métricos constantes que afectan al número de sílabas \cite{quilis1984metrica}: 
	
	\begin{itemize}
		\item Cuando un verso es oxítono, se cuenta una sílaba más. 
		\item Cuando un verso es paroxítono, se cuenta las sílabas reales existentes
		\item Cuando un verso es proparoxítono, se cuenta una sílaba menos. 
	\end{itemize}
	
	Además de estas denominaciones, la tradición poética ha clasificado los versos en función de su patrón métrico. Un verso cuenta con un número finito de acentos. Un endecasílabo, por ejemplo, puede contar con cinco acentos rítmicos o menos; un octosílabo hasta cuatro. Por lo tanto, el número de combinaciones posibles también son finitas y asentadas en la tradición poética. Estos esquemas métricos tradicionales son los que el poeta tiene en mente cuando elige un metro. En el presente trabajo, se toma la clasificación de \namecite{pou2020metrica}, por ser la más sistemática y exhaustiva. En la Tabla \ref{table:clasificacion-endecas} se recoge la tipología del endecasílabo.
	
	\begin{table}[]
		\begin{tabular}{@{}cll@{}}
			\toprule
			\multicolumn{1}{l}{Endecasílabos} & \multicolumn{1}{c}{} & \multicolumn{1}{c}{} \\ \midrule
			\multicolumn{1}{l}{— Heroicos}    & 2.6.10               & puro                 \\
			& 2.4.6.8.10           & pleno                \\
			& 2.4.6.10             & corto                \\
			& 2.6.8.10             & largo                \\
			& 2.4.10               & difuso               \\
			\multicolumn{1}{l}{— Melódicos}   & 3.6.10               & puro                 \\
			& 1.3.6.8.10           & pleno                \\
			& 3.6.8.10             & largo                \\
			& 1.3.6.10             & corto                \\
			\multicolumn{1}{l}{— Sáficos}     & 4.8.10               & puro                 \\
			& 1.4.8.10             & puro pleno           \\
			& 1.4.6.8.10           & pleno                \\
			& 4.6.10               & corto                \\
			& 1.4.6.10             & corto pleno          \\
			& 4.6.8.10             & largo                \\
			& 2.4.8.10             & largo pleno          \\
			& 4.10                 & difuso               \\
			& 1.4.10               & difuso pleno         \\
			\multicolumn{1}{l}{— Dactílicos}  & 4.7.10               & puro                 \\
			& 1.4.7.10             & pleno                \\
			& 2.4.7.10             & corto                \\
			\multicolumn{1}{l}{— Enfáticos}   & 1.6.10               & puro                 \\
			& 1.6.8.10             & pleno                \\
			\multicolumn{1}{l}{— Vacíos}      & 6.10                 & puro                 \\
			& 6.8.10               & pleno                \\
			& 1.4.10               & pleno                \\
			& 2.4.10               & heroico              \\ \bottomrule
		\end{tabular}
		\label{table:clasificacion-endecas}
		\caption{Clasificación del verso de once sílabas \cite[p. 118]{pou2020metrica}.}
	\end{table}
	No obstante, en la realización de un verso, la distribución de los acentos puede ser variada y no coincidir exactamente con los ofrecidos en la clasificación. En el Ejemplo \ref{ej:gongora}, se da el patrón métrico 1.6.(7).10, el cual es posible y frecuente en el endecasílabo, pero no se observa en la Tabla \ref{table:clasificacion-endecas}. Se advierte, sin embargo, que tiene los acentos característicos en 1.6.10 del ritmo enfático puro, siendo el acento en séptima extrarrítmico. 
	\begin{subexamples}
		\textit{Siempre la claridad viene del cielo}\\
		\textit{\textbf{Siem}-pre-la-cla-ri-\textbf{dad}-\textbf{vie}-ne-del-\textbf{cie}-lo} \\
		11 | 1.6.(7).10 \\
		(Claudio Rodríguez) \\
		\label{ej:gongora}
	\end{subexamples}
	
	Siguiendo este criterio, cada verso se puede clasificar en función de la proximidad a cada uno de los tipos: se observan los acentos característicos del tipo de verso y se tratan los no coincidentes como extrarrítmicos.
	
	Cuanto mayor es el número de acentos, más complejo y rico es el verso. Sin embargo, cuando supera las once sílabas, se rompe en semiversos o hemistiquios. Por ejemplo, un verso de 14 sílabas, también llamado alejandrino, se suele dividir en hemistiquios de 7+7. Es decir, el verso en español tiene el límite rítmico del endecasílabo.
	
	Precisamente, por esa subdivisión en hemistiquios, se puede realizar una escansión separada de cada uno de los semiversos. Así, en el Ejemplo \ref{ej:juanramon} se contempla que el primer hemistiquio es proparoxítono, y resta una sílaba al cómputo. El segundo es oxítono y suma una. El resultado final es un verso alejandrino. 
	
	\begin{subexamples}
		\textit{Oh, qué frescor, qué música / de chopos de estación} \\
		14 = 8 - 1 / 6 + 1 \\
		(Juan Ramón Jiménez) \\
		\label{ej:juanramon}
	\end{subexamples}

	La silabación es la segmentación de una palabra en sus sílabas constituyentes. Esta tarea viene determinada por la estructura morfológica del español \cite[1.3.4a]{espanola2009nueva} 
	
	Es importante reparar en que la tarea de silabación es distinta de la tarea de contar sílabas. Por ejemplo, dada la palabra \textit{bolígrafo}, la silabación resultaría \textit{bo-lí-gra-fo}, donde se ha hecho un esfuerzo morfológico por establecer la relación entre grafemas de la palabra. No obstante, el cómputo sílabico no informa de la relación entre los grafemas, sino que es una cuantificación: 4. 
	
	A parte de estas consideraciones generales, los versos son sometidos por los poetas a varios recursos métricos que afectan al cómputo silábico e introducen ambigüedad en la determinación del patrón rítmico. Los más habituales son la sinalefa, dialefa, sinéresis y diéresis. 
	
	Sinalefa es la pronunciación en una sola sílaba de dos o más vocales que se encuentran en palabras distintas (Ejemplo \ref{ej:amalia}). 
	
	\begin{subexamples}\label{ej:amalia}
		\textit{Creía que te había dicho adiós} \\
		\textit{Cre-\textbf{í}-a-que-t\textbottomtiebar{eha}-\textbf{bí}-a-\textbf{di}-ch\textbottomtiebar{oa}-\textbf{diós}} \\
		11 | 2.6.8.10
		
		(Amalia Bautista)
	\end{subexamples}
	
	En los versos de arte mayor superiores a once sílabas, la pausa hemistiquial rompe la sinalefa. En el Ejemplo \ref{ej:machado} se rompe la sinalefa entre \textit{mente} y \textit{entre}.
	
	\begin{subexamples}\label{ej:machado}
		\textit{Escucho solamente entre las voces una} \\
		\textit{Es-\textbf{cu}-cho-\textbf{so}-la-\textbf{men}-te-entre-las-\textbf{vo}-ces-\textbf{u}-na} \\
		7 + 7 = 14 \\
		
	\end{subexamples}

	La dialefa es la ruptura en dos sílabas de dos vocales de palabras distintas, es decir, lo contrario de la sinalefa. 
	
	La sinéresis consiste en la pronunciación en una sola sílaba de dos o más vocales que normalmente forman dos sílabas. ``Hermosas ninfas que en el río metidas'' (Garcilaso): ``Hermosas ninfas que en el \textbf{rio} metidas'' (11) \cite[Capítulo III]{pou2020metrica}
	
	Por último, la diéresis consiste en la pronunciación de dos o más vocales que normalmente se realizan como un diptongo, convirtiéndolo en un hiato. Aunque suele venir explicitada por el poeta con el signo ¨. No siempre es así. Sirva de muestra el verso del Ejemplo \ref{ej:herrera}, donde se encuentra una diéresis no indicada.
	
	\begin{subexamples}
		\textit{Todas las tardes se muere un niño} \\
		\textit{\textbf{to}-das-las-\textbf{tar}-des-se-\textbf{mü}-e-\textbf{\textbottomtiebar{reun}}-\textbf{ni}-ño} \\
		11 | 1.4.8.(9).10\\
		
		(Federico García Lorca) \\
		\label{ej:herrera}
	\end{subexamples}
	
	Además, la resolución de ambigüedades no es exacta. En el verso del Ejemplo \ref{example.4}, existen dos posibles escasiones, ambas válidas. En cualquier caso, de acuerdo con lo dicho anteriormente, se podría argumentar que el esquema 2.4.(7).8.10 con acento extrarrítmico en 7ª es natural del endecasílabo sáfico largo pleno y es, con un grado alto de certidumbre, el que el autor trataba de conseguir; por el contrario, el patrón 3.7.8.10 es artificial y su sonoridad está cuestionablemente alejada del endecasílabo. Hacia este tipo de decisiones se orienta el sistema presentado.

	\begin{subexamples}[preamble={\textit{Juez Elisio, que de un verde probo}}]\label{example.4}
		\item\label{example.4a}
		\textit{Ju-\textbf{ez}-E-\textbf{li}-sio-que-\textbf{d\textbottomtiebar{eun}}-\textbf{ver}-de-\textbf{pro}-bo} \\
		11 | 2.4.(7).8.10\\
		\item\label{example.4b}
		\textit{\textbf{Juez}-E-\textbf{li}-sio-que-de-\textbf{un}-\textbf{ver}-de-\textbf{pro}-bo} \\
		11 | 3.7.8.10 \\
		(Lope de Vega)
	\end{subexamples}

	\section{Trabajos relacionados}
	\label{sec:relacionados}
	
	La métrica, dentro de la filología, es un tema fundamental en el análisis literario de un poema. Entre los trabajos clásicos en el campo se encuentran los manuales de \namecite{bello1859principios}, \namecite{tomas1956metrica} y \namecite{quilis1984metrica}. Más recientemente se publicó el diccionario y manual de Domínguez Caparrós  \cite{caparros1993metrica,caparros2001diccionario}. El presente trabajo se apoya en las últimas investigaciones de métrica española expuestas por \namecite{pou2020metrica}

	En cuanto a los algoritmos de escansión automática, en los últimos años se encuentra el trabajo de \namecite{navarro2017metrical}. Se trata de un sistema basado en reglas que emplea el analizador morfológico Freeling \cite{padro2012freeling}. Se centra en la resolución de ambigüedades; marca sinalefas y diéresis, pero no trata las sinéresis. Procesa las ambigüedades empleando una base de conocimiento de los distintos patrones métricos extraída en un análisis estadístico; aunque ofrece información adicional sobre el corpus, este análisis no es estrictamente necesario para realizar escansión, pues lo patrones vienen ya dados por la propia preceptiva de la métrica. Además, el sistema se evalúa sólo sobre versos de métrica fija. 
	
	Por otro lado, \namecite{agirrezabal2017comparison} entrenan una red de neuronas bidireccional LSTM a nivel de carácter. El sistema predice patrones métricos a partir de una transformación enriquecida de la entrada, que incluye silabación. La salida no informa sobre la decisión del sistema, por lo que no se pueden conocer las ambigüedades que ha detectado.
	
	El sistema más reciente publicado es Rantanplan \cite{de2020rantanplan}. Se trata de un método basado en reglas que emplea un sistema de silabación del 99.99\% de precisión. La señalización de acentos se apoya en esta silabación y en el etiquetado gramatical de la librería Spacy \cite{honnibal2017spacy}. Como indican los autores, el sistema está limitado por este modelo estadístico: comete errores ocasionales asignando los acentos y, desde el punto de vista del coste computacional, tan sólo su carga retrasa 18 segundos el análisis de acuerdo con las pruebas de los investigadores. No tienen en cuenta los efectos de compensación y dialefa hemistiquiales, lo cual perjudica la tasa de acierto (\textit{accuracy}) sobre el corpus de poemas polimétricos.

	Todos los métodos anteriores utilizan de una manera u otra métodos de silabación. Tampoco tienen en cuenta los fenómenos métricos de los versos de más de once sílabas.

	\section{\textit{Jumper}: algoritmo de análisis métrico sin silabación}
	\label{sec:propuesta}
	En esta Sección se presenta \textit{Jumper}, el método de análisis métrico automático desarrollado. 
	Se divide a su vez en tres subsecciones: análisis de palabra, verso y poema.
	
	\subsection{Módulo 1: Análisis de palabra: cómputo de sílabas y acentos}
	\label{sec:computo-palabra}
	Como se ha dicho en la Introducción, en español, el núcleo de una sílaba es siempre vocálico. La unidad de la sílaba viene dada por la vocal. Por lo tanto, para contar sílabas de manera eficiente, basta con contar las vocales de una palabra, teniendo en cuenta los diptongos. De hecho, no es necesario localizar los hiatos explícitamente: los diptongos en español son un conjunto finito de 14 realizaciones. Por lo tanto, dada la aparición contigua de dos vocales, basta comprobar si es un diptongo para no contar esa siguiente vocal como sílaba. 
	
	Además, este mismo procedimiento permite localizar el acento. Se asume que la palabra sigue las reglas de acentuación gráfica del español \cite[3.4]{espanola2010ortografia}. Así, mientras contamos vocales, se guarda la posición en la que se ha encontrado una vocal acentuada gráficamente. Si tras el cómputo no se ha encontrado el acento, se aplican las reglas de acentuación básicas del español para hallarlo.
	
	En cuanto se conoce si la palabra es aguda, llana o esdrújula, se puede calcular fácilmente la compensación cuando se encuentra al final del verso.
	
	Gracias a este método, al tener en cuenta la vocal acentuada explícitamente, se puede también localizar el acento cuando el poeta emplea el recurso de la sístole.\footnote{Sístole consiste en adelantar una o más sílabas el acento normal de una palabra; jilguero - jílguero. \cite[Capítulo III]{pou2020metrica}}
	
	Cabe decir que, de momento, se está operando a nivel de palabra. En cuanto se traten de resolver las ambigüedades métricas de un verso producidas por las sinéresis sí será necesario dar un tratamiento específico a los hiatos.
	

	\subsection{Módulo 2: Análisis de verso}
	
	Una vez se ha diseñado la función que devuelve el número de sílabas y el lugar de sus acentos individualmente, se quiere extender el análisis al verso entero. Este módulo es el núcleo del método, y es el que entraña más complejidad. Se divide, a su vez, de tres submódulos. El primero se encarga del cómputo de sílabas y acentos; el segundo, de la compensación hemistiquial para los versos mayores de once sílabas y, el tercero, de resolución de ambigüedades métricas.
	
	\subsubsection{Submódulo 1: Cómputo de sílabas y acentos de un verso}
	
	El primer submódulo funciona de la siguiente manera. Dado un verso, se convierte en una lista de palabras. Para cada palabra de la lista, se calculan, con la función expuesta en la Sección \ref{sec:computo-palabra}, el número de sílabas, el lugar del acento en esas sílabas y el factor de compensación. Este factor es -1 para las palabras esdrújulas, 0 para las llanas y 1 para las agudas. Se suma el número de sílabas de la palabra al número total de sílabas del verso, teniendo en cuenta, si la hubiera, la sinalefa con la palabra siguiente. Se calcula la posición del acento y se añade a la lista del patrón métrico. Si la palabra es un adverbio en \textit{-mente} se tienen en cuenta sus dos acentos. Si la palabra es átona, no se añade a la lista de acentos.
	
	En este primer submódulo se calcula la medida del verso teniendo en cuenta todas las sinalefas. Posteriormente, si hay alguna que no habría que tener en cuenta, se resuelve en el módulo de ambigüedades.
	
	\subsubsection{Submódulo 2: Compensación hemistiquial}
	
	Al comienzo del análisis de un verso, no se conoce cuántas sílabas va a tener. Si tras el cómputo del primer submódulo resulta un verso de más de once sílabas, se hace una llamada recursiva sobre la función indicando que ahora se tenga en cuenta la compensación hemistiquial, activando así el segundo submódulo. Si durante este segundo cómputo se hallan hemistiquios, se tiene en cuenta el factor de compensación de cada semiverso y la ruptura de la sinalefa por la pausa hemistiquial, si la hubiera. Se considera que una palabra se encuentra en la frontera de un hemistiquio cuando el número de sílabas del verso en proceso sumado al número de sílabas de la palabra y su factor de compensación resultan la medida esperada del hemistiquio (por ejemplo, 7 para el alejandrino). 
	
	\subsubsection{Submódulo 3: Resolución de ambigüedades}
	
	Para calcular correctamente el patrón métrico de un verso, hay que tener en cuenta los recursos métricos empleados en él. Es necesario detectar dialefas, sinéresis y diéresis, sus posibles combinaciones, y obtener el mejor candidato.
	
	Por lo tanto, el submódulo de resolución de ambigüedades tiene dos tareas. La primera consiste en detectar las posibles ambigüedades del verso y generar candidatos con ellas. La segunda consiste en elegir, entre los posibles candidatos, el mejor.
	
	La activación del submódulo de detección de ambigüedades se indica desde el módulo superior de análisis de poema. Una vez indicada, por eficiencia, la detección se hace al mismo tiempo que el cómputo del Submódulo 1. Cuando se detecta una sinalefa, se genera a la vez un candidato con dialefa. Cuando se detecta un diptongo, se genera un candidato con diéresis. Cuando se detecta un hiato, se genera un candidato con sinéresis. Una vez se han almacenado todos los candidatos con las posibles ambigüedades, se combinan: por ejemplo, un verso puede presentar, al mismo tiempo, una dialefa y una sinéresis. Todos los candidatos se etiquetan con su recurso métrico para así conservar la decisión del submódulo de resolución de ambigüedades.
	
	En la segunda tarea, se elige el mejor candidato. Como se ha expuesto en la introducción, los patrones métricos de los versos son finitos y sus posiciones se aproximan a los tipos de versos asentados en la tradición. Por lo tanto, se compara la proximidad del vector del patrón rítmico extraído con cada uno de los tipos asentados en la preceptiva literaria; en función de esa proximidad, se elige el mejor candidato.
	
	Aunque probar todas las posibles ambigüedades podría parecer costoso computacionalmente, es rápido por dos razones. La primera es que los versos ambiguos son sólo un pequeño subconjunto del corpus. La segunda es que, gracias al sencillo algoritmo de cómputo de sílabas y acentos, el coste de volver a computar un verso es muy bajo.
	
	\subsection{Módulo 3: Análisis de poema}
	
	Dado un poema, se convierte en una lista de versos. Se realiza la escansión de cada uno de ellos por medio del módulo de análisis a nivel de verso, con el sistema de resolución de ambigüedades desactivado.
	
	Una vez calculadas todas las medidas y patrones rítmicos, se observa la frecuencia de los números de sílabas. Pueden darse dos casos. El primero, cuando sólo hay una medida frecuente: en este caso, se concluye que es un corpus de metro fijo. Si la lista de medidas frecuentes es mayor de uno (por ejemplo, en la poesía contemporánea, son habituales los poemas de versos polimétricos de 7, 11, y 14 sílabas), se concluye que es un corpus de poemas polimétricos. Una vez se tiene la lista de medidas frecuentes, se vuelven a computar los versos que no coincidan con ellas, ahora con el módulo de resolución de ambigüedades activado. La única diferencia entre poesía de metro fijo y mixto es que, para esta última, la lista de medidas frecuentes se actualiza en un contexto de un número $n$ de versos. El tamaño del contexto es el único hiperparámetro del sistema. Por defecto, se establece en 14, ya que el número de versos del poema por antonomasia, el soneto.
	
\begin{figure*}
		\includegraphics[width=\textwidth,height=4cm]{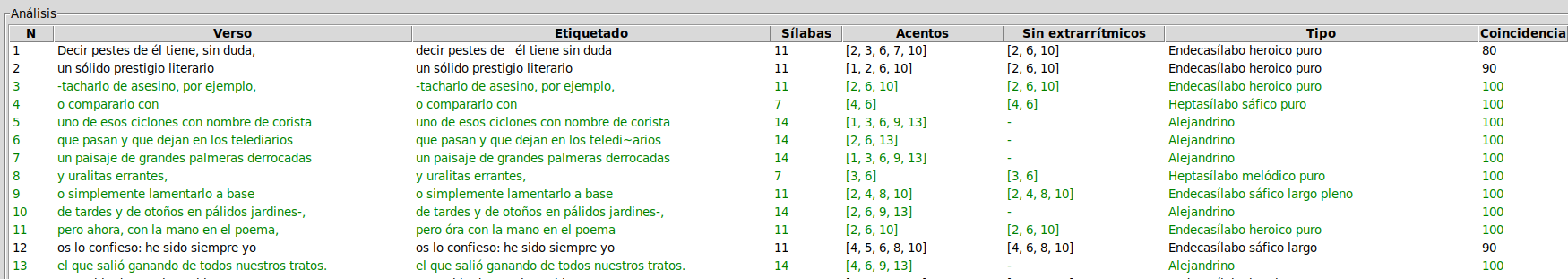}
		\caption{Salida del sistema. Observése la diéresis detectada en el décimo verso.}
		\label{fig:analisis-app}
	\end{figure*}

	Una vez resueltas las ambigüedades, se almacena en una tabla cuyas columnas son: verso, verso etiquetado, número de sílabas, patrón métrico en forma de vector de enteros, patrón métrico en forma de vector de enteros del tipo de verso sin acentos extrarrítmicos, nombre del verso y, por último, ratio de coincidencia entre el patrón calculado y el patrón sin acentos extrarrítmicos. La salida del sistema se muestra en la Figura \ref{fig:analisis-app}. 
	
	\section{Evaluación}
	\label{sec:evaluación}
	En los últimos años se ha estandarizado el uso del corpus anotado de sonetos (estrofa de 14 versos de once sílabas) del Siglo de Oro \cite{navarro2016metrical} para la evaluación de los sistemas de escansión sobre poemas de medida fija. Originalmente, consistía en un conjunto de 1400 versos etiquetados manualmente con su patrón métrico; posteriormente se ampliaron a 10268. Está disponible en la herramienta de descarga de corpus Averell del proyecto POSTDATA.\footnote{https://github.com/linhd-postdata/averell}
	
	Asimismo, para métrica de medida mixta, \namecite{de2020rantanplan} introdujeron recientemente un corpus elaborado a partir de la antología de Antonio \namecite{carvajal1983extravagante}. Se compone de 4378 versos, tanto de arte mayor como de arte menor, etiquetados manualmente. Por razones de derechos de autor, no es de dominio público; no obstante, sus desarrolladores nos permitieron su uso para esta experimentación.
	
	Los sistemas se evalúan en función de su acierto (\textit{accuracy}) establecido de forma binaria para cada verso: el acierto es 1 si se han identificado correctamente todos los acentos anotados del verso, y 0 en caso contrario. El rendimiento del sistema se mide como la tasa de acierto sobre el total de versos del corpus.
	
	Todos los experimentos de nuestro sistema se han realizado sobre un ordenador equipado con la misma configuración que se empleó para las evaluaciones de los distintos métodos comparados en Rantanplan: procesador Intel\textsuperscript{®} Core™ i7-8550U CPU @ 1.80GHz y 16GiB de memoria RAM DDR4. 
	
	El acierto (\textit{accuracy}) se reporta en el intervalo [0,1] con dos cifras significativas. El tiempo se indica en segundos.
	
    Al sistema se le ha dado el nombre de \textit{Jumper}.
	
	\subsection{Evaluación sobre poemas de medida fija}
	
	La Tabla \ref{table.3} recoge la tasa de acierto (\textit{accuracy}) y el tiempo de cada uno de los métodos sobre el corpus de 10268 versos de medida fija elaborado por \namecite{navarro2016metrical}. Nuestro sistema, Jumper, obtiene una tasa de acierto de 0,95, lo que supone una mejora del 2,2\% respecto al actual estado del arte establecido por Rantanplan. Además, se ejecuta 21 veces más rápido que éste.
	
	Estos resultados indican que prescindir de silabación supone una mejora en el enfoque de la solución al problema. Gracias a que no se requiere la compilación de numerosas expresiones regulares para la separación de sílabas ni los pesados modelos de PoS-Tagging para la asignación de acentos, el tiempo es menor. La velocidad conseguida permite realizar una desambiguación exhaustiva en poco tiempo, lo que influye también en la mejora de la tasa de acierto.
	
	\begin{table} [htbp]
		\begin{center}
			\begin{tabular} {lcr}
				\hline\rule{-2pt}{15pt}
				{\bf Método} & {\bf \textit{Accuracy}} & {\bf Tiempo}\\
				\hline\rule{-4pt}{10pt}
				Navarro-Colorado & 0,91 & 16787s \\
				Rantanplan & 0,93 &53s\\
				Jumper (nuestro) & \textbf{0,95} & \textbf{2,5s}\\
				\hline
			\end{tabular}
		\end{center}
		\caption{\label{table.3} \textit{Accuracy} y tiempo sobre el corpus de 10268 versos de métrica fija de Navarro-Colorado. Los resultados de Navarro-Colorado y Rantanplan se han extraído de \namecite{de2020rantanplan}. Los de Jumper se han obtenido bajo la misma configuración de hardware.}
	\end{table}
	
	Es interesante examinar el resultado de la evaluación sobre el subconjunto inicial de 1400 versos establecido por sus autores, ya que es el único corpus para esta tarea sobre el que se ha reportado una \textit{inter-annotator agreement} (IAA) o tasa de acuerdo entre anotadores. La IAA sobre este subconjunto es de 0,96 \cite{navarro2016metrical}. Jumper obtiene un 0,95. No obstante, se ha detectado en el análisis de errores de la Sección \ref{sec:errores-fija} un sesgo en la anotación de estos primeros 1400 versos: la interjección ``oh", frecuente en el Siglo de Oro, es considerada átona sólo en este subconjunto del corpus. Así, únicamente añadiendo a la lista de palabras átonas esa interjección Jumper obtendría un 0,96. Por tanto, nuestro sistema alcanza el límite de precisión que se puede medir con este corpus, ya que iguala la tasa de acuerdo entre anotadores. Este conjunto también permite comparar el efecto del tiempo de arranque de los sistemas sobre conjuntos de datos de tamaño pequeño. Bajo la misma configuración, Jumper analiza los 1400 versos en 0,33 segundos frente a los 2356 del sistema de Navarro-Colorado y los 21 segundos de Rantanplan.
	
	\subsection{Evaluación sobre poemas de medida mixta}
	
	En la Tabla \ref{table.4} se contempla la tasa de acierto (\textit{accuracy}) sobre el corpus de métrica mixta. Jumper obtiene 0,82, lo que supone un 25\% de mejora relativa respecto al actual estado del arte establecido por Rantanplan. En cuanto al tiempo, se ejecuta 25 veces más rápido que éste.
	
		\begin{table} [htbp]
		\begin{center}
			\begin{tabular} {lcr}
				\hline\rule{-2pt}{15pt}
				{\bf Método} & {\bf \textit{Accuracy}} & {\bf Tiempo}\\
				\hline\rule{-4pt}{10pt}
				Navarro-Colorado & 0,49 & 7484s \\
				Rantanplan & 0,65 & 27s \\
				Jumper (nuestro) & \textbf{0,82} & \textbf{1,1s} \\
				\hline
			\end{tabular}
		\end{center}
		\caption{\label{table.4}\textit{Accuracy} y tiempo de ejecución sobre el corpus de 4300 versos de métrica mixta de Antonio Carvajal.  Los resultados de Navarro-Colorado y Rantanplan se han extraído de \namecite{de2020rantanplan}, donde ambos han sido ejecutados en el mismo entorno. Los de Jumper se han obtenido bajo la misma configuración de hardware.}
	\end{table}
	
	Estos resultados son producto de los beneficios de prescindir de silabación y librerías de etiquetado gramatical, ya comentados en la Sección anterior. Pero, además, se añade otro factor determinante en la mejora sobre este corpus: la consideración de los fenómenos métricos particulares de los versos de más de once sílabas. Esta consideración permite clasificar sus patrones métricos con mayor acierto. En el corpus de medida mixta de Carvajal se encuentran mezclados tanto versos de 7, 9 u 11 como de 14. El Ejemplo \ref{ej:carvajal}, tomado del conjunto de datos, Jumper lo clasifica correctamente como verso de 14 sílabas con ritmo  1.3.6.8.10.13, ya que el primer hemistiquio es oxítono y suma una sílaba. Rantanplan, sin embargo, lo clasifica como verso tridecasílabo de ritmo 1, 3, 6, 7, 9, 12, al no tener en cuenta la compensación. Lo mismo ocurre con el fenómeno de ruptura de sinalefa entre hemistiquios.
	
	\begin{subexamples}[preamble={\textit{una lucha común, y un descanso común}}]\label{ej:carvajal}
		\item\label{ej:carvajala}
					6+1 / 6+1 = 14 | 1.3.6.8.10.13   (Jumper)   \\
		\item\label{ej:carvajalb}
				13 | 1.3.6.7.9.12  (Rantanplan)\\
		(Antonio Carvajal)
	\end{subexamples}
	
	\subsection{Análisis de errores}
	
	En esta Sección se profundiza en las circunstancias en las que Jumper falla en su clasificación de patrones métricos. En el corpus de medida fija se clasifican incorrectamente 546 de 10268 patrones. Sobre el corpus de medida mixta se clasifican incorrectamente 800 de 4378. 
	
	\subsubsection{Errores sobre medida fija}
	\label{sec:errores-fija}
	Se ha tomado una muestra de 100 versos de los 546 clasificados incorrectamente. Tras su análisis manual, se contempla que hay principalmente cuatro fuentes de error. 
	
	La primera fuente de error (35 equivocaciones), atribuible al sistema, es la elección incorrecta del mejor candidato entre las posibles realizaciones de un verso ambiguo. Cuando un verso tiene dos escansiones posibles igual de correctas, ambas con el mismo ratio de coincidencia con el patrón asentado en la tradición, la resolución de este empate no siempre se hace correctamente. Dado el verso del Ejemplo \ref{example:hernando}, el sistema devuelve el ritmo 3.6.8.10, que es igual de correcto que el anotado, 2.6.8.10. Sin embargo, cuando el acento en la sexta sílaba recae en un diptongo ``cruel'', la diéresis ``crü-el'' es más natural que la dialefa entre ``si'' y ``a'' elegida por Jumper.
	
	\begin{subexamples}[preamble={\textit{si a Silvia la cruel pastora viere}}]\label{example:hernando}
		\item\label{example:hernandoa}
		\textit{si-a-\textbf{Sil}-via-la-\textbf{cruel}-pas-\textbf{to}-ra-\textbf{vie}-re,}\\
		11	| 3.6.8.10 (Jumper)\\
		\item\label{example:hernandob}
		\textit{s\textbottomtiebar{ia}-\textbf{Sil}-via-la-crü-\textbf{el}-pas-\textbf{to}-ra-\textbf{vie}-re} \\
		11 | 2.6.8.10 (Anotado)\\
		(Hernando de Acuña)
	\end{subexamples}	
	
	La segunda fuente de error es la anotación manual incorrecta, que se detecta en 26 de los 100 versos analizados. En algunas ocasiones, los errores se deben a desambiguaciones imprecisas del anotador. El Ejemplo \ref{example:lope1} se ha anotado con el ritmo 2.3.7.10, el cual es extraño al endecasílabo. La escansión 2.(3).6.10 del endecasílabo heroico puro es la correcta para ese verso.
	
	\begin{subexamples}[preamble={\textit{dolor pide a Felipe de Liaño}}]\label{example:lope1}
		\item\label{example:lope1a}
		\textit{do-\textbf{lor}-\textbf{pi}-d\textbottomtiebar{ea}-Fe-\textbf{li}-pe-de-Lï-\textbf{a}-ño}\\
		11 | 2.3.6.10 (Jumper)\\
		\item\label{example:lope1b}
		\textit{do-\textbf{lor}-\textbf{pi}-de-a-Fe-\textbf{li}-pe-de-\textbf{Lia}-ño} \\
		11 | 2.3.7.10 (Anotado)\\
		(Lope de Vega)
	\end{subexamples}
	
	La tercera fuente de error (21 equivocaciones), también relacionada con la anotación, se debe a la consideración de palabras átonas como tónicas y viceversa. Por ejemplo, en los primeros 1400 versos se suele considerar átona la interjección ``oh'', mientras que, en los restantes, se considera tónica. Jumper la considera tónica. Solo debido a la arbitrariedad de la acentuación de esta interjección, tan frecuente en el Siglo de Oro, se detectan en el análisis manual 11 errores. 
	
    La cuarta fuente de error son 8 versos con errores ortográficos que provocan asignaciones de acentos incorrectas.
    
    La atribución de los 10 errores restantes es ambigua.

	\subsubsection{Errores sobre medida mixta}
	
	Se ha revisado manualmente una muestra de 100 de los 800 versos clasificados incorrectamente por Jumper.
	
	Se han hallado 21 errores de anotación. Es importante recalcar que el corpus de Carvajal ha sido recientemente introducido para esta tarea y se encuentra todavía en un proceso de refinamiento. De hecho, la herramienta puede contribuir a limpiar la anotación, puesto que si se revisan los errores de Jumper se encuentra que uno de cada cinco casos es atribuible a un error manual de anotación y no a un error del sistema.
	
	Otra fuente de error es que en ocasiones los anotadores no consideran tónicos algunos determinantes indefinidos y demostrativos: es el caso de ``un, una, unos, unas, este, esta...''. La mayor parte de las veces, estos determinantes se encuentran en acentos extrarrítmicos, de modo que se anulan por el acento rítmico contiguo. Sin embargo, es un acento que realmente se encuentra en el verso, y aunque posteriormente en un análisis abstracto se elimine, consideramos que se ha de tener en cuenta en la escansión. Como Jumper, de acuerdo con \namecite[9.7b]{espanola2009nuevafonetica}, sí los considera acentuados, falla en 46 de los 100 versos analizados por esta razón. Sin embargo, como hemos explicado, podría considerarse también un error de anotación. En nuestras pruebas comprobamos que si se añaden estos determinantes tónicos a la lista de palabra átonas, el acierto sobre este corpus baja del 0,82 al 0,76, de lo que se deduce que los anotadores en algunas ocasiones consideran estos determinantes átonos y en otras ocasiones tónicos.
	
	En menor medida (6 casos), se han hallado errores derivados de no emplear etiquetado gramatical. Por ejemplo, cuando se procesa la palabra ``mientras'' nuestro sistema la considera átona, ya que su función más frecuente es conjuntiva (``Estudia \textit{mientras} yo leo'') \cite[p. 25]{quilis1984metrica}; sin embargo, cuando tiene función de adverbio, es tónica ``estudia; \textit{mientras}, yo leo''. Este problema no es una dificultad insalvable con el sistema implementado, ya que se pueden resolver este tipo de errores tratándolos como una ambigüedad más. En cualquier caso, la frecuencia tan baja de errores atribuibles a la falta de etiquetado gramatical confirma que la estrategia de Jumper es preferible. 
	
	Del mismo modo que en el corpus de medida fija, el sistema falla 23 veces por no elegir correctamente el mejor candidato entre los versos ambiguos.
	
	Los 4 casos restantes se deben a errores ortográficos que provocaban asignaciones de acentos incorrectas.
	
	\section{Interfaz gráfica para Jumper}
	\label{sec:app}
	Se ha desarrollado una aplicación de escritorio para ofrecer una interfaz al método expuesto. Mientras el usuario escribe el poema, se ejecuta el algoritmo y se imprime el análisis en tiempo real. Para cada verso, si no tiene acentos extrarrítmicos se imprime en verde. Si no coincide plenamente con el esquema acentual sin extrarrítmicos, se imprime en negro. Si no cumple la tendencia versal del poema, en rojo. La tendencia versal se calcula de forma automática si el usuario no la introduce explícitamente. Esta aplicación puede ser de utilidad para investigadores de la poesía española y traductores. Puede descargarse en el siguiente enlace: \texttt{https://github.com/grmarco/jumper}
	
	\section{Conclusiones}
	\label{sec:conclusiones}
	En este trabajo se ha presentado un algoritmo de escansión automática de poemas en español sin necesidad de silabación. A partir del conocimiento de la métrica española, se han establecido una serie de premisas que han simplificado el problema. De manera resumida, las premisas son:
	
	\begin{itemize}
		\item La tarea de silabación es distinta a la de contar sílabas.
		\item La vocal es núcleo de la sílaba en español; por lo tanto, es lo que identifica su unidad. Así, se ha podido desarrollar un sencillo algoritmo de cómputo de sílabas y acentos (Sección \ref{sec:computo-palabra}), que ha mejorado notablemente la eficiencia sin necesidad de uso de librerías externas. 
		\item Ante un verso ambiguo, las realizaciones posibles del patrón métrico son finitas y asentadas en la tradición. Por lo tanto, se trata de resolver las ambigüedades teniendo en cuenta la aproximación a los patrones naturales del verso.
		\item La compensación y dialefa hemistiquial son determinantes para realizar escansiones precisas. 
	\end{itemize}
	
	Gracias a estas premisas, se ha desarrollado un sistema análisis métrico automático que simplifica el problema de la medida del verso, tiene en cuenta la compensación hemistiquial y resuelve las ambigüedades métricas derivadas sinalefas, dialefas, sinéresis y diéresis, sin perder precisión ni información sobre la decisión del analizador. 
	
	Nuestro algoritmo, Jumper, mejora el actual estado del arte en un 2\% para la clasificación de patrones métricos sobre poesía de medida fija, y en un 25\% sobre poesía de medida mixta. Además, todas las evaluaciones se han ejecutado entre 21 y 25 veces más rápido que el estado del arte. 
	
	También se ha llevado a cabo un análisis de los errores cometidos por el sistema, lo que permite vertebrar trabajos futuros para solucionarlos, y también depurar los corpus utilizados, ya que entre 1/5 y 1/4 de los supuestos errores del sistema son en realidad problemas de la anotación manual. 
	
	Finalmente, se ha desarrollado una interfaz gráfica para el algoritmo de análisis métrico en tiempo real, que puede ser de utilidad para investigadores de poesía española.
	
	\section*{Agradecimientos}
	Esta investigación se ha desarrollado gracias al proyecto MISMIS-BIAS (PGC2018-096212-B-C32), financiado por el Gobierno de España, Ministerio de Ciencia, Innovación y Universidades. 
	
	Para la evaluación del sistema expuesto han sido de ayuda la facilitación de los corpus empleados por parte de los autores de Rantanplan \cite{de2020rantanplan}.
	
	\bibliographystyle{fullname}
	\bibliography{marco_scansion_sepln_2020}
	
	\appendix
	\section{Apéndice: Código fuente y reproducibilidad}
	
	Se pueden reproducir los resultados del analizador en este repositorio: \texttt{https://github.com/grmarco/jumper-evaluation}
	
\end{document}